\pdfoutput=1

\documentclass[11pt]{article}

\usepackage{ACL2023}

\usepackage{times}
\usepackage{latexsym}
\usepackage{booktabs}
\usepackage{dingbat}
\usepackage{amsmath}

\usepackage[T1]{fontenc}

\usepackage[utf8]{inputenc}

\usepackage{microtype}

\usepackage{inconsolata}

\usepackage{graphicx} 
\usepackage{enumitem}
\usepackage{multirow}
\usepackage{algorithm}
\usepackage{algorithmic}
\usepackage{xcolor} 
\usepackage{array}

\PassOptionsToPackage{table}{xcolor}

\usepackage{caption}
\usepackage{subcaption}

\definecolor{lightgray}{gray}{0.9}

\usepackage{array}

\usepackage{xcolor}
\usepackage{xspace}

\newcommand{\vlm}{\textsc{vlm}\xspace}
\newcommand{\vicuna}{\textsc{vicuna}\xspace}
\newcommand{\vqa}{\textsc{vqa}\xspace}
\newcommand{\gqa}{\textsc{gqa}\xspace}
\newcommand{\clevr}{\textsc{clevr}\xspace}

\newcommand{\llm}{\textsc{llm}\xspace}

\newcommand{\gpt}{\textsc{gpt-4v}\xspace}

\newcommand{\ours}{\textsc{Logic2Vision}\xspace}
\newcommand{\ourdata}{\textsc{VisReas}\xspace}

\title{\textsc{VisReas}: Complex Visual Reasoning with Unanswerable Questions}

\author{Syeda Nahida Akter$^{1}$, Sangwu Lee$^{2}$, Yingshan Chang$^{1}$, Yonatan Bisk$^{1}$, Eric Nyberg$^{1}$\\
Language Technologies Institute, Carnegie Mellon University, Pittsburgh, PA, United States$^{1}$ \\
Department of Computer Science, University of Rochester, Rochester, NY, United States$^{2}$ \\
\texttt{\{sakter,yingshac,ybisk,ehn\}@cs.cmu.edu}\ , \ \texttt{slee232@u.rochester.edu}}

\usepackage{bibentry}

\begin{document}

\maketitle
\begin{abstract}



Verifying a question's validity before answering is crucial in real-world applications, where users may provide imperfect instructions. In this scenario, an ideal model should address the discrepancies in the query and convey them to the users rather than generating the best possible answer. Addressing this requirement, we introduce a new compositional visual question-answering dataset, \textbf{\ourdata}, that consists of answerable and unanswerable visual queries formulated by traversing and perturbing commonalities and differences among objects, attributes, and relations. \textbf{\ourdata} contains 2.07M semantically diverse queries generated automatically using Visual Genome scene graphs. The unique feature of this task, \textit{\textbf{validating question answerability with respect to an image before answering}}, and the poor performance of state-of-the-art models inspired the design of a new modular baseline, \textbf{\ours} that reasons by producing and executing pseudocode \textit{\textbf{without any external modules}} to generate the answer. \ours outperforms generative models in \ourdata (+4.82\% over LLaVA-1.5; +12.23\% over InstructBLIP) and achieves a significant gain in performance against the classification models.

\end{abstract}



\section{Introduction}\label{sec:intro}

\begin{figure}[ht!]
    \centering
    \includegraphics[width=\linewidth]{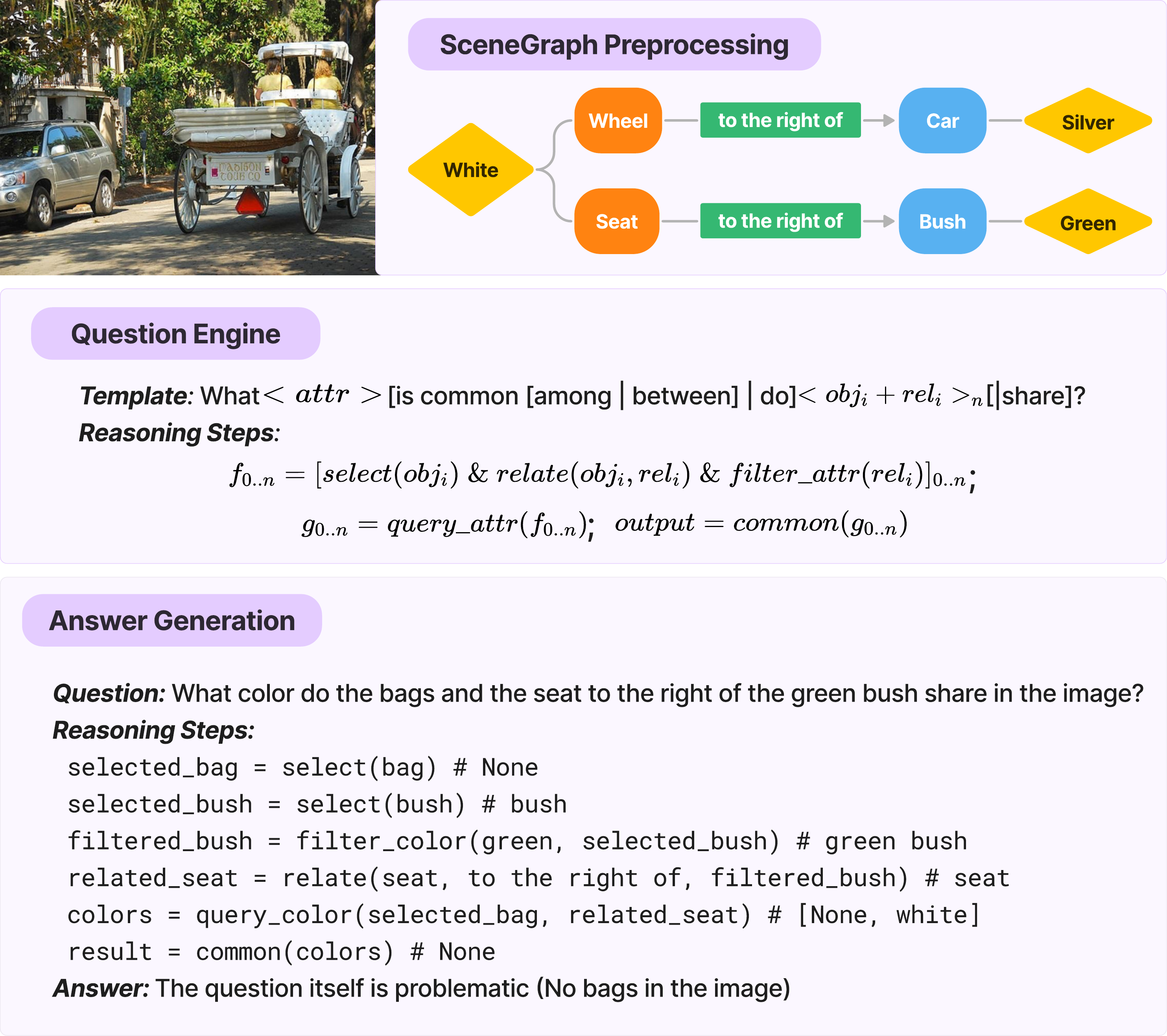}
    \caption{Overview of \textsc{VisReas} dataset construction process. Using scene graphs, we cluster objects (\textcolor{orange}{\textbf{orange}}), relations, and attributes of the related objects (\textcolor{blue}{\textbf{blue}}) based on the attribute of the corresponding objects (\textcolor{orange}{\textbf{orange}}). Then the question engine takes each template as input and traverses all possible clusters to generate the query as well as the reasoning steps. Each function in the reasoning steps can return \textbf{\textsc{NONE}} if any object, attribute, or relation is absent in the image.}
    \label{fig:visreas}
\end{figure}

In visual question answering (\vqa), validating question authenticity with the corresponding image and then reasoning over it is an important requirement in real-world application dynamics where users may make errors in judgment, leading to invalid queries. Confirming a question’s validity becomes pivotal to maintaining consistency, rectifying mistakes, and preventing misguided responses \cite{rajpurkar-etal-2018-know}. Following the prior \vqa datasets’ \cite{balanced_vqa_v2, krishnavisualgenome, hudson2018gqa} focus on answerable questions only, a system trained solely for answerable questions may exhibit unstable behaviors when faced with unanswerable queries. For instance, a delivery robot receiving an incorrect address but a valid instruction like \textit{\texttt{“place the package by the yellow door”}} might overlook the error unless prompted to reevaluate its decision. In contrast, presuming the correctness of the query would likely lead to unpredictable behaviors. Therefore, a reliable and responsible system should be able to question the validity of the instruction it receives before acting upon it. 


However, aligning questions with the region of interest in the image 
breaks down visual reasoning task into 
perception (object detection and scene representation learning) and reasoning (question interpretation and inference grounded in the scene). Datasets and models proposed to date have shown significant improvement in the detection task which therefore improved the perception system \cite{balanced_vqa_v2, krishnavisualgenome, tan2019lxmert}, but they face critical vulnerabilities due to the lack of generalities in the datasets \cite{balanced_binary_vqa, agrawal-etal-2016-analyzing}. 
Recent datasets \cite{johnson2017clevr, selvaraju2020squinting, hudson2018gqa} 
encourage reasoning beyond surface-level object recognition and focus on 
multi-step inference. But they tend to reason about object relations \textit{(often questions revolving around single object)} instead of reasoning over clusters of objects in the image that share common attributes or relations. Reasoning over general sets of objects requires both identifying objects and understanding their attributes and relations.
Where prior scene-graph based work assumes reasoning follows from traversing a single path to generate an answer, our goal is to establish a multi-hop approach of identifying 
cliques with shared properties.


%

Bridging the gap in prior benchmarks, we introduce a new dataset, \textbf{\textsc{VisReas}} (\textbf{Vis}ual \textbf{Reas}oning), for studying 
reasoning over commonalities and differences across objects. 
The unnatural assumption in the current VQA datasets - \textit{``a correct answer for every question''} causes models to produce an answer even when the question is inapplicable and has no possible answer. To ensure that models verify the consistency of question text with the image before answering, we curate questions that have no answer given the image by altering relations and attributes among the objects. We design a question generation engine that takes the information about objects, attributes, and relations from 
the Visual Genome scene graphs \cite{krishnavisualgenome} and finds common features shared among multiple objects. Based on this retrieved information, we 
generate $2.07$M unique questions covering vast semantic variations. Each question is paired with a scene graph and a semantic program that specifies the series of reasoning steps needed to be performed to produce the answer. 
Our generated questions require visual reasoning abilities such as comparing, differentiating, counting, clustering objects, and performing logical reasoning. Most importantly, unlike other \vqa datasets, \textsc{VisReas} enforces the \vqa models to verify the information in the question with the image in each reasoning step before predicting an answer.


We find existing \vqa models 
less robust in the reasoning and unanswerable settings presented by \ourdata. 
Motivated by the shortcomings of existing models, we propose a new architecture, \textbf{\ours} that has been trained to produce logical reasoning steps from the query at first and then predict answers based on the reasoning steps and the image. Unlike prior generative models, \ours is compute and cost-efficient as it does not require any external expensive APIs or modules and solely relies on the reasoning capabilities of visual language models (\vlm). Experiments on \textsc{VisReas} shows that \ours outperforms the current fine-tuned \vqa models: obtaining \textbf{66.20\%} (+4.82\% over LLaVA-1.5 \cite{liu2023improvedllava}, +12.23\% over InstructBLIP \cite{dai2023instructblip}) accuracy on \textsc{VisReas}.


In short, our contributions are twofold: 
\begin{itemize}[leftmargin=*]
    \item We introduce 
    \textbf{\ourdata}, a dataset
    containing complex yet natural reasoning. Our dataset makes the first step towards developing reliable \vlm adaptable to real-world scenarios where user instructions may not always be impeccable. 
    \item We present 
    \textbf{\ours}, that aims to handle spatial reasoning by executing consecutive pseudocode with verification in each step.  
\end{itemize}
\section{Related Works}

Recent years have witnessed tremendous progress in visual understanding. Multiple attempts have been made to mitigate the systematic biases of VQA datasets \cite{balanced_vqa_v2, zhang2016yin, agrawal2018don, johnson2017clevr}, but they fall short in providing an adequate solution: Some approaches operate over constrained and synthetic images \cite{zhang2016yin, johnson2017clevr}, neglecting the realism and diversity natural photos provide. 
\citet{suhr-etal-2019-corpus} introduced a dataset for reasoning about semantically-diverse natural language descriptions of images in the form of a 
classification task. While the dataset exhibits diverse semantic phenomena, 
this task rarely requires much beyond a single type of object recognition and its associated relation and attribute. 
Unlike these datasets, 
\ourdata is open-ended and consists of both unanswerable and answerable queries based on the similarity/dissimilarity of multiple objects in the image. \ourdata jointly evaluates \vqa models’ alignment, multihop reasoning, and verification ability which cannot be approximated by simply finding the most likely object/relation/attribute to answer the question.
 

Recent transformer-based models have 
\cite{tan2019lxmert, Lu_2020_CVPR, nguyen2022coarse} achieved promising performance on visual reasoning tasks. Yet, these models are prone to reproducing spurious correlations without accurately learning true causal relations \cite{agrawal-etal-2016-analyzing, jia-liang-2017-adversarial, 10.5555/3237383.3237389}. 
Neural-symbolic methods \cite{andreas2016neural, hu2017learning, hudson2018compositional, hudson2019learning} explicitly perform symbolic reasoning on the object and language representations. These models offer modularity and interpretability in the reasoning process. 
However, as module parameters are usually derived solely from end-task supervision, there is a potential for the program to deviate from accurately explaining the model's behavior \cite{10.5555/3172077.3172259, jain-wallace-2019-attention, subramanian-etal-2020-obtaining}.

Conversely, a recent 
approach to modularity
leverages
Large Language Models (\llm) to craft code or Python programs using expensive APIs 
\cite{chen2021evaluating, surismenon2023vipergpt, Gupta_2023_CVPR, subramanian-etal-2023-modular}. 
However, these approaches outsource basic aspects of the reasoning to external components rather than performing reasoning as part of the model itself.
For example, prior works outsource basic cognitive abilities such as recognizing objects, counting, and even arithmetic operations. Focusing on these limitations, our proposed \ours aims to leverage single {\vlm} to address complex reasoning in a modular approach that shows promising performance across models of three different categories.


\section{\ourdata: Visual Reasoning}


The \ourdata dataset is an attempt towards 
better aligning model capabilities with real application circumstances. In parallel, \ourdata aims to develop complex compositional reasoning into the machine that involves consideration of relations among multiple objects and verification of alignment between information provided in the question and the image. 
In the following sections, we provide details about the \ourdata data generation pipeline and a comprehensive analysis of the \ourdata dataset. In the supplementary material, we conduct a detailed comparative study between \ourdata and the well-established \gqa dataset, followed by details of the human evaluation process using Mechanical Turk. 

\subsection{Data Generation}

Our dataset is constructed in three major steps: (1) Process scene graphs, (2) Define templates and reasoning functions that the question will involve, (3) Automatically generate corresponding reasoning steps in pseudocodes along with the final answer from each query as shown in Fig. \ref{fig:visreas}. Finally, to prevent models from learning statistical biases in attribute, reasoning, or answer type distributions, we meticulously balance the \ourdata dataset across three distinct paradigms (\autoref{appendix: data_balance}). 

\subsubsection{Scene Graph Processing}  To begin with the data construction process, we run two phases of processing on the scene graphs before passing them to the question engine.

\textbf{First Phase.} We clean up the scene graphs by 
removing opposite attributes and discarding object nodes with similar names that share similar attributes and relations. Our processed scene graphs contain 1703 distinct objects, 14 attributes, and 114 relationships. It is also observed that one object name in the image might correspond to multiple object IDs and bounding boxes in the scene graph. This will cause ambiguity in the later question-generation process. Thus, we merge bounding boxes corresponding to the same object name with a high IoU ($>$ 0.7). In addition, there can be images where a bigger bounding box contains multiple small bounding boxes, which can be either parts of the object represented by the bigger bounding box (e.g., a cat (bigger bounding box) has a tail, ear, face (small bounding boxes), etc.) or they can collectively represent the object in the bigger bounding box (e.g., lime and apple can together be mentioned as fruits). These overlapping bounding boxes will be problematic while clustering objects based on similar attributes (e.g., fruits and lime are all green; for \textit{`What has the same color as the lime?'} the answer generation module will produce: fruits and apple - which is ambiguous). To discard these cases, we measure the ratio of intersection area vs individual bounding box area and check whether the smaller objects are subclasses of the bigger one using Wordnet \cite{miller-1994-wordnet}.  If the ratio is high and the larger object is a superclass of the smaller one, we discard the larger bounding box during preprocessing to avoid ambiguity.

\textbf{Second Phase.} We cluster the scene graphs based on the common attributes and relations among the objects in each image and create several sub-graphs as seeds for the question engine. Initially, we cluster objects based on a single relation or attribute, later we merge the clusters recursively if there are objects with multiple attributes or relations in common. Finally, each cluster represents a collection of objects that share a similar set of attributes and relations and the question engine exhaustively traverses all clusters to generate questions. For each object in a cluster, we also store other objects that are related to that object along with their relation name. This information is used to populate nested compositional references for multi-hop relation traversal.

\subsubsection{Question Engine} For question generation from the clusters, we manually create 182 templates on different attributes (\autoref{tab:ques_stats}). Our templates cover five categories of reasoning (\textit{query}, \textit{count}, \textit{compare}, \textit{verify}, and \textit{choose}) which can be further broken down into nine broad categories of reasoning mentioned in 
Appendix. For some categories, we have list answers and no-answer cases. All of our templates are formulated considering clusters of objects 
to facilitate multi-object comparison. To generate no-answer cases, we apply two approaches: (1) We either add an outlier (object not present in the image) to the cluster or include an object that exists in the image but not in the cluster and has different relations and attributes from the objects in the cluster. (2) We perturb the existing relation/attribute of an object inside a cluster (e.g., change \textit{`apple to the left of knife'} to \textit{`apple to the right of knife'}) which derives no-answer cases.

\begin{table}[]
  \begin{center}
  \begin{small}
  \scalebox{0.95}{
  \begin{tabular}{lccc}
  \toprule
    \textbf{Attribute} & \textbf{Templates} & \textbf{Train}      & \textbf{Validation}\\\toprule
    Color	       &          12	       & 1326086	&  1500\\
    Cleanliness	   &          8	       & 7794	    &  900\\
    Material	   &          15	       & 368337	    &  1500\\
    Size	       &          4	       & 116438	    &  1500\\
    Pose	       &          18	       & 36687	    &  1500\\
    Height	       &          10        & 9894	    & 1200\\
    Weather	       &          6	       & 31376	    & 1500\\
    Length	       &          11	       & 45764	    & 1500\\
    Tone	       &          11	       & 37184	    & 1500\\
    Shape	       &          15	   & 30119	    & 1500\\
    Activity	   &          21	   & 15639	    & 1500\\
    Sport Activity &          21	   & 13215	    & 1500\\
    Age	           &          12	   & 19594	    & 1500\\
    Pattern	       &          18	   & 14313	    & 1500\\\hline
     \textbf{Total}	       &          \textbf{182}	   & \textbf{2072440}	& \textbf{20100}\\\toprule
  \end{tabular}}
  \end{small}
  \end{center}
 \caption{Question-template distribution over attributes}
\label{tab:ques_stats}
\end{table}

\subsubsection{Answer Generation} The answer generation step involves two consecutive phases. \textit{Initially,} we formulate the reasoning steps in pseudocode (\autoref{fig:visreas}) and produce the intermediate results for each line of code using our designed parser (\autoref{fig: semantic_parser}). For each question template and reasoning type, we have hand-coded the basic reasoning steps necessary to answer the query. Based on the number of objects, relations, and attributes, our parser generates all intermediate reasoning steps along with the answers. \textit{Finally,} we combine all intermediate results to come up with the answer. If any intermediate reasoning step results in \textbf{`NONE'}, the final answer becomes \texttt{`the question itself is problematic'} indicating some objects, relations, or attributes mentioned in the question text cannot be found in the image.

\begin{table}[]
\centering
\resizebox{\columnwidth}{!}{%
\begin{footnotesize}
\begin{tabular}{@{}l@{\hspace{10pt}}c@{\hspace{10pt}}c@{\hspace{10pt}}cc@{\hspace{2pt}}c@{\hspace{2pt}}c@{\hspace{2pt}}c@{\hspace{2pt}}c@{\hspace{2pt}}c@{\hspace{2pt}}c@{}}
                & \textbf{\# QA}     & \textbf{\# Images} & \textbf{AveQLen} & 
                \rotatebox{90}{\scriptsize \textbf{ListAns}} 
                & \rotatebox{90}{\scriptsize \textbf{Grounding}} 
                & \rotatebox{90}{\scriptsize \textbf{Counting}} 
                & \rotatebox{90}{\scriptsize \textbf{RealWorld}} 
                & \rotatebox{90}{\scriptsize \textbf{No Answer}}\\ 
\toprule 
\textbf{VQA}             & 614,163   & 204,721   & 6.2 $\pm$ 2.0    & \checkmark  &           &       \checkmark     & \checkmark        & \\ 
\textbf{Visual7W}        & 327,939   & 47,300    & 6.9 $\pm$ 2.4    & \checkmark  & \checkmark         & \checkmark         & \checkmark         & \\ 
\textbf{CLEVR}           &       853,554    &         100,000  &    18.3$\pm$ 3.5        &      &  \checkmark  &      \checkmark     &           &        &   \\ 
\textbf{GQA}             & 1,750,623 & 113,000   &     7.9 $\pm$ 3.1             &    &      \checkmark     &           &      \checkmark     & \\ 
\midrule
\textbf{\ourdata} & 2,072,437 & 113,000   &      19.4 $\pm$ 4.6       &  \checkmark  &     \checkmark      &      \checkmark     &   \checkmark & \checkmark        \\ 
\bottomrule
\end{tabular}%
\end{footnotesize}
}
\caption{Comparisons on existing VQA datasets. \ourdata covers a wide variety of reasoning along with \textbf{\textit{No Answer}} cases. The average question length is also higher in \ourdata compared to others.}
\end{table}

\subsection{Dataset Analysis and Comparison}
\begin{figure}[h!]
    \centering
    \includegraphics[width=\columnwidth]{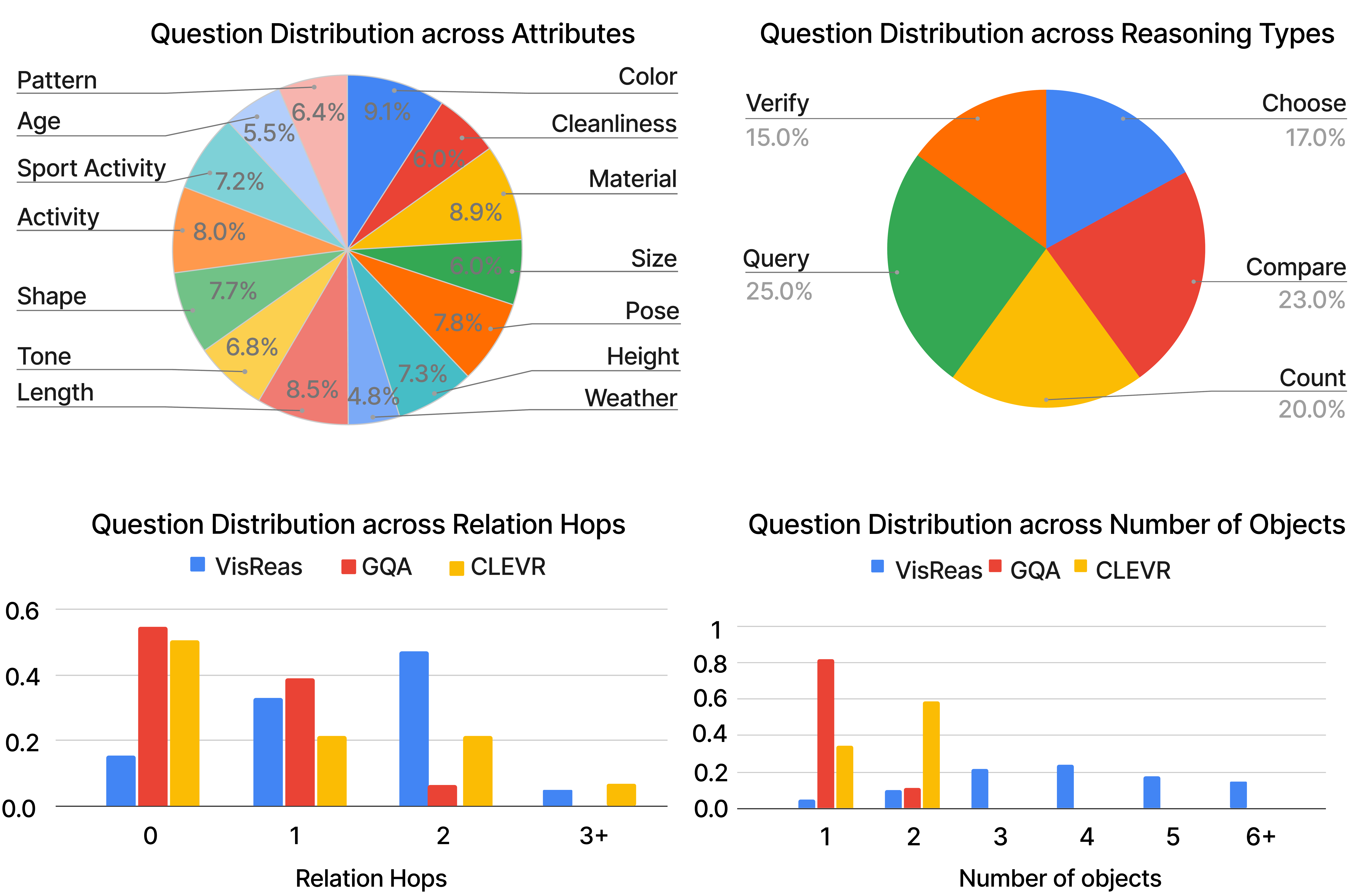}
    \caption{Overview of \ourdata statistics. \textbf{(Top left)} The dataset covers 14 attributes in a balanced ratio. \textbf{(Top right)} It consists of five reasoning types of queries in a balanced distribution. \textbf{(Bottom left)} Comparison of multi-hop relation traversal for different VQA datasets. Majority questions of \ourdata require multi-hop traversal compared to others. \textbf{(Bottom right)} Comparison of number of objects mentioned in the question for different datasets where \ourdata questions contain larger amount of objects.}
    \label{fig: question_stats}
\end{figure}

The \ourdata dataset consists of 113K images from the Visual Genome where each image is annotated with dense descriptions of the scene stored in the scene graphs. We refine the existing scene graphs and generate $2,072,437$M unique questions, twice the size of current \vqa datasets (\autoref{tab:ques_stats}), that combine features of multiple objects and their relations and require the implementation of consecutive complex reasoning skills with an in-depth understanding of object attributes and relations in the image. Our dataset covers 14 different attributes and 114 diverse relations among 1703 different objects from real-life images. We define five major types of reasoning (\autoref{fig: question_stats}) while generating the corpus based on the overall nature of the query template. 
\autoref{fig:ques_type} shows details of the query structures along with examples. However, the intermediate reasoning steps that are necessary to answer the query can be diverse and can combine all five types of reasoning for a single query (as in \autoref{fig:visreas}). We balance the dataset combinedly across 14 attributes and 5 reasoning types (\autoref{appendix: data_balance}).

Compared to existing \vqa tasks, \ourdata emphasizes creating longer reasoning chains (multi-hop) with a larger number of objects (\autoref{fig: question_stats}). The average number of reasoning hops for \ourdata is $1.42$ ($95$\% CI: $[1.415,1.417]$), significantly higher than \gqa (mean: $0.52$; $95$\% CI: $[0.517,0.519]$) and \clevr (mean: $0.84$; $95$\% CI: $[0.839,0.843]$).  However, to limit the question length and increase human readability (\autoref{fig: readability})
, the majority of the questions require at most two hops relation traversal for each object.



Reflecting on human clustering ability based on commonalities, 
\ourdata consists of queries that require consideration of multiple objects based on their attribute or relation similarities. Therefore, unlike existing datasets, the majority of \ourdata queries are composed of more than three objects from the image. The average objects per question for \ourdata is $3.91$, which is higher than both \gqa ($1.12$) and \clevr ($1.63$). Hence, \ourdata requires multiple object detection and consecutive reasoning to answer a single query (\autoref{fig: question_stats}). In addition, each query can have multiple attributes associated with it (\autoref{fig: ques_n_attributes}). 
For example, in question, \texttt{`What is the common material among the silver and blue utensils?'}, both \texttt{<material>} and \texttt{<color>} attributes are needed to be considered for answer generation that involves multiple attribute filtering along with the associated objects. 
\begin{figure}[ht!]
    \centering
    \includegraphics[width=\columnwidth]{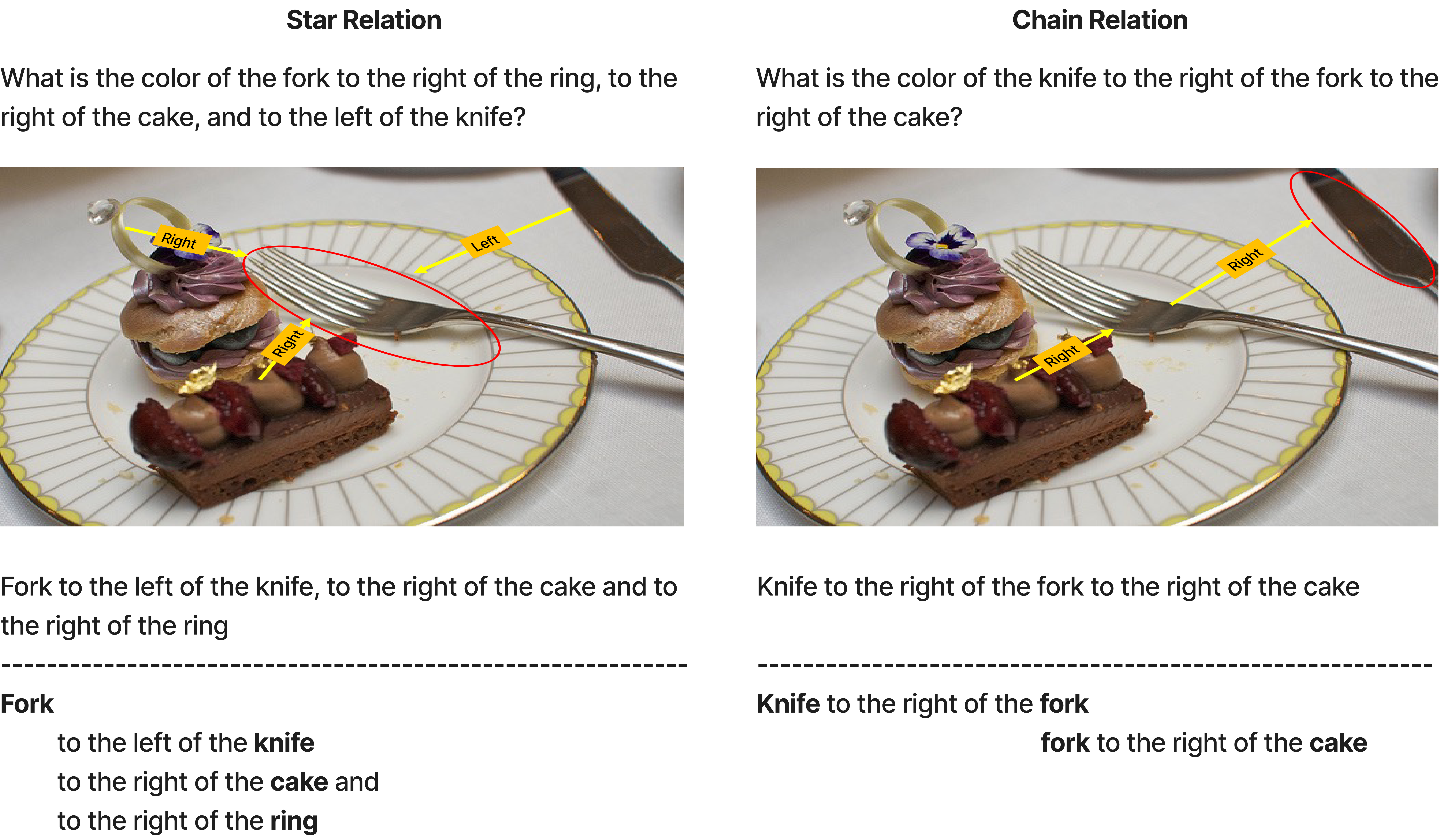}
    \caption{\ourdata contains two types of relation traversals. \textbf{Star relation} states a single object that shares multiple relations with other objects (Left). \textbf{Chain relation} states multiple objects that share a single relation with each other (Right).}
    \label{fig: relations}
\end{figure}

In contrast to other spatial reasoning datasets that focus primarily on one-hop relation traversals (Bottom left of \autoref{fig: question_stats}), we explore two ways of novel traversals: (1) \textbf{\textit{Star Relation:}} The target object shares multiple relations with other objects (e.g. is the center of the star and other objects are connected to it with a relation -- \autoref{fig: relations} left), and (2) \textbf{\textit{Chain Relation}}: The target object is related to an object that is related to another object and the relation traversal is linear (\autoref{fig: relations} right). The inclusion of these traversals adds multi-hop complexity to the corpus and makes the \textit{each-step verification process} harder for unanswerable questions (as \autoref{fig: star-chain-code}). 





\section{\ours}

\begin{figure*}[h!]
    \centering
    \includegraphics[width=\textwidth]{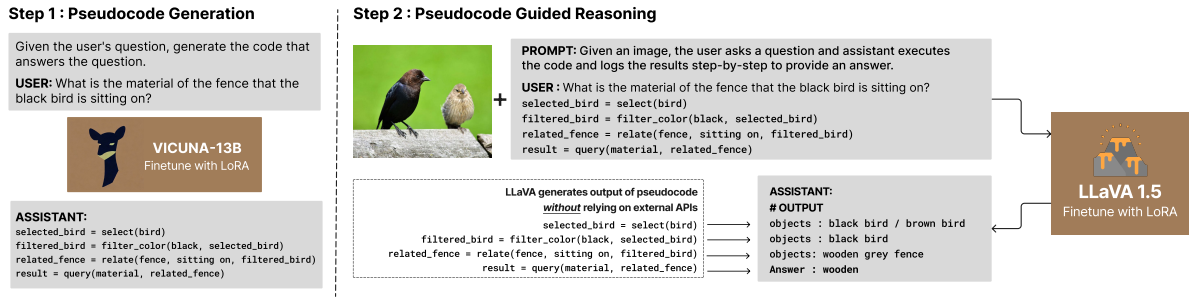}
    \caption{Overview of \textbf{\ours}. In \textbf{Pseudocode Generation} phase, we generate pseudocode which outlines the reasoning steps. 
    During \textbf{Pseudocode-Guided Reasoning}, the pseudocodes along with the question and image are provided to the model. The model executes all intermediate pseudocodes to arrive at the final answer.}
    \label{code2vision}
\end{figure*}


In recent years, {\llm}s combined with code generation and chain-of-thought prompting have shown impressive performance in complex reasoning by generating intermediate reasoning steps before inferring the answer \cite{zhang2023multimodal, surismenon2023vipergpt}. 
However, these frameworks are often prone to hallucinations of {\llm}s and are too restricted in terms of reasoning they can perform and dependent on expensive external modules to execute the reasoning \cite{zhang2023multicot, surismenon2023vipergpt}. To address these limitations and elicit the reasoning capability of {\vlm}s,  we propose \ours, a two-stage \vqa framework that (1) plans the necessary reasoning steps using the question and (2) executes the plan with the help of an image leveraging the SOTA \vlm (\autoref{code2vision}).

\subsection{Stage 1: Pseudocode Generation}

Given a natural language question, this module generates a consecutive set of reasoning steps as pseudocodes. For training our pseudocode generation model, we take advantage of the existing \vqa dataset: \gqa 
as it provides a semantic string that decomposes the question into a sequence of reasoning steps. For instance, the semantic string for the question \texttt{`Is there a red apple on the table?'} would be \texttt{`select: table $\rightarrow$ relate: on, subject, apple $\rightarrow$ exist: ?'}. We build a custom parser (\autoref{fig: semantic_parser}) that converts each line of \gqa semantic string to pseudocode and extract the expected output from the scene graph. The parsed \texttt{(pseudocode, output)} pairs serve as a rationale to solve the question (\autoref{fig: pseudocode}). For the pseudocode generation model, we use an instruction finetuned \vicuna-13B \cite{vicuna2023} model which has shown good performance across various language tasks including code generation. We finetune \vicuna using LoRA \cite{hu2022lora} to generate the pseudocode for given a question. 
The finetuned model achieves 98.6\% METEOR \cite{banerjee-lavie-2005-meteor} score and 96.3\% ROGUE-L \cite{lin-2004-rouge} score against ground-truth code parsed from \gqa semantic strings.

\subsection{Stage 2: Pseudocode-Guided Reasoning}

Since the Pseudocode Generation module outlines the necessary steps to answer the question, the remaining task is to perform pseudocode-guided sequential reasoning on the image. For this stage, we choose state-of-the-art \vlm, LLaVA-1.5 \cite{liu2023improvedllava}, due to its impressive performance in diverse reasoning tasks. As LLaVA-1.5 was not trained to reason with pseudocode and image, we fine-tuned it to generate an answer by executing sequential reasoning with the pseudocode and the image. To adapt this framework in our case, we rearrange the instruction as: 


\noindent\fbox{%
\centering
    \parbox{0.99\columnwidth}{%
\noindent{\small\textcolor{blue}{<Image>} \textit{Given an image, the user asks a question and assistant executes the code and logs the results step-by-step to provide an answer.} \textbf{Question}: \{\textcolor{purple}{Question}\} \textbf{Code}: \{\textcolor{teal}{Code}\} \textbf{Outputs}: \{\textcolor{brown}{Outputs}\} \textbf{Answer}:}
}%
}\newline

The essential training details of this stage can be found in \autoref{subsec: code2vision}.



\section{Experiments and Analysis}

In the subsequent sections, we conduct a comprehensive analysis of the \ourdata dataset and assess the performance of various benchmarks including \ours, \gpt \cite{openai2023gpt4}, and human participants, revealing a notable disparity from human performance. 



\subsection{Baseline Experiments}

To analyze the complexity and generalizability of our dataset and model, we run experiments with models trained on both classification and generative tasks.  We cover two types of generative models: \textbf{\textcolor{blue}{GEN}} (relies on pretrained visual-language alingment module) and \textbf{\textcolor{purple}{Code-GEN}} (generates a program and utilizes external APIs to solve \vqa tasks). We categorize \ours as \textbf{\textcolor{teal}{Logic-GEN}} as it produces intermediate logical reasoning steps before answering. All model configurations can be found in \autoref{appendix: baseline_config}. To make the training and inference consistent, we define our own prompt for all generative models (as \autoref{subsec: gen_configs}). Table \ref{tab: baselines} shows the results of different baselines on both \gqa and \ourdata. \autoref{tab: baselines_breakdown} shows the finetuned results on \ourdata and further breaks down the performance of each model in each reasoning type along with \gpt and human accuracy.

\begin{table}[]
\centering
\resizebox{\columnwidth}{!}{%
\begin{footnotesize}
\begin{tabular}{lllcc}
\multirow{2}{*}{} & \multirow{2}{*}{}                       & \multicolumn{1}{c}{\multirow{2}{*}{\textbf{Model}}} & \multicolumn{2}{c}{\textbf{Accuracy (\%)}}                                                \\ 
                                  &      & \multicolumn{1}{c}{}                         & \multicolumn{1}{c}{\textbf{\gqa}} & \multicolumn{1}{c}{\textbf{\ourdata}} \\
\toprule
 \multirow{6}{*}{\textbf{\rotatebox{90}{ZS}}} & \multirow{3}{*}{\textbf{\textcolor{blue}{GEN}}}           & BLIP-2    \shortcite{li2023blip2}                                           & \multicolumn{1}{c}{44.70}         & \multicolumn{1}{c}{35.16}   \\ 
                           &             & InstructBLIP   \shortcite{dai2023instructblip}                                    & \multicolumn{1}{c}{49.50}  & \multicolumn{1}{c}{36.84}       \\
                        &                & LLaVA-1.5   \shortcite{liu2023improvedllava}                                    & \multicolumn{1}{c}{63.3$^{*}$}  &  \multicolumn{1}{c}{38.98}         \\\cmidrule{2-5}
& \multirow{2}{*}{\textbf{\textcolor{purple}{Code-GEN}}}      & ViperGPT   \shortcite{surismenon2023vipergpt}                                          & \multicolumn{1}{c}{48.10}          & \multicolumn{1}{c}{10.31}         \\ 
                              &          & VisProg  \shortcite{Gupta_2023_CVPR}                                            & \multicolumn{1}{c}{50.50}    & \multicolumn{1}{c}{20.82}         \\ \midrule\midrule
\multirow{5}{*}{\textbf{\rotatebox{90}{FT}}} & \multirow{3}{*}{\textbf{\textcolor{orange}{CLS}}}           & LXMERT   \shortcite{tan2019lxmert}                                            & 60.05                         &    50.15  \\ 
                          &              & ViLBERT    \shortcite{Lu_2020_CVPR}                                          & 60.65                         & 53.05   \\ 
                          &              & CRF    \shortcite{nguyen2022coarse}                                            & 72.10                          &    53.56  \\ \cmidrule{2-5}
& \textbf{\textcolor{teal}{Logic-GEN}} & \ours                                           & \textbf{60.32}                        &  \textbf{66.20}    \\ \bottomrule
\end{tabular}%
\end{footnotesize}
}
\caption{Performance comparison among baseline models on GQA and \textsc{VisReas}. (*) GQA trainset images were used during training.}
\label{tab: baselines}
\end{table}

\textbf{[\textcolor{orange}{CLS}] For models trained with classification task,} we finetune and evaluate on both \gqa and \ourdata. From the fine-tuning results of the CLS models, it is obvious that \ourdata proposes a different task than \gqa that can not be easily solved by scaling the model size or changing the pretraining corpus.  Furthermore, the higher performance gap of the models between \gqa and \ourdata tasks suggests the inefficacy of the existing CLS models on our proposed spatial reasoning task.



\begin{table*}[h!]
\centering
\resizebox{2\columnwidth}{!}{%
\begin{tabular}{l|ccc|ccc|cc|c|c|c}
\multirow{2}{*}{\textbf{Metric}} & \multicolumn{3}{c|}{\textbf{\textcolor{orange}{CLS}}}                                                                 & \multicolumn{3}{c|}{\textbf{\textcolor{blue}{GEN}}}                      & \multicolumn{2}{c|}{\textbf{\textcolor{purple}{Code-GEN}}}                                                  & \textbf{\textcolor{teal}{Logic-GEN}}    & \multirow{2}{*}{\textbf{GPT-4V}} &\multirow{2}{*}{\textbf{Humans}} \\ 
                                 & LXMERT    & ViLBERT  & CRF          & BLIP-2         & InstructBLIP & LLaVA-1.5  & ViperGPT  & VisProg  & \ours  &                                  \\ 
                                 \toprule
Choose                           & 74.23     & 82.91    &    83.30     & 71.21          & 78.50 &    84.11    & 10.37     & 15.86    & 82.54                 & 82.61 & 91.30       \\ 
Compare                          & 65.62     & 69.86    &    71.87     & 28.72          & 53.29       &  67.75 & \phantom{0}5.97      & 26.09    & 59.25                 & 68.33 & 86.12      \\ 
Count                            & 45.32     & 47.80    &    49.59     & 25.88          & 49.86       &  43.08 & \phantom{0}7.85      & \phantom{0}6.02     & 39.47                 & 39.52 & 85.78     \\ 
Query                            & 44.05     & 47.65    &    48.11     & 41.55         & 47.77       & 50.31 & \phantom{0}4.35      & 19.30    & 63.79                 & 58.78 & 81.78        \\
Verify                           & 76.10     & 82.18    &    83.03     & 70.77          & 49.48     & 81.27   & \phantom{0}3.10      & 44.18    & 84.54                 & 82.16 & 93.94        \\ 
Problematic                      & 67.54     & 77.08    &    78.41     & 25.39           & 64.68     &  68.04   & \phantom{0}0.00      & \phantom{0}0.00     & 55.34                 & 70.18 & 90.29         \\ 
Non-Problematic                  & 56.11     & 59.16    &    61.60     & 51.41          & 52.25      &  60.31 & 11.97     & 24.17    & 67.94                 & 55.47 & 84.89              \\ 
\midrule
\textbf{Accuracy (\%)}           & \textbf{50.15} & \textbf{53.05} & \textbf{53.56}        & \textbf{47.81} & \textbf{53.97}        & \textbf{61.38} &\textbf{10.31} & \textbf{20.82} & \textbf{66.20}        & \textbf{62.83} & \textbf{87.21}                   \\ 
\bottomrule
\end{tabular}%
}
\caption{Accuracy breakdown of baseline models and humans on \textsc{VisReas} across different reasoning types. \textbf{Problematic} type consists of questions that contain certain relation, attribute, or object that is missing/ not consistent with the image. In contrast, \textbf{Non-Problematic} questions have correct answers as the question is consistent with the image. Except for the \textcolor{purple}{Code-GEN} models, we provide \textbf{fine-tuned results} on \textsc{VisReas} for all other models.}
\label{tab: baselines_breakdown}
\end{table*}

\textbf{[\textcolor{blue}{GEN}] From generative domain,} we select three SOTA models, BLIP-2, InstructBLIP, and LLaVA-1.5, that try to leverage the LLMs using two types of vision-language alignment modules: Q-Former and MLP cross-modal connector. 
We evaluate the models on zero-shot \gqa and \ourdata to probe the relevance of our proposed task to their training domain. We notice that BLIP-2 performs poorly on our task compared to \gqa where InstructBLIP and LLaVA-1.5 shows higher accuracy. 
Both LLaVA-1.5 and InstructBLIP are instruction tuned on diverse downstream tasks which allows them to excel in \vqa tasks compared to BLIP-2. However, LLaVA-1.5 gains the highest zero-shot accuracy in this category due to its training set images being overlapped with \ourdata. Yet, it shows a significant drop (-24.32\%) in ZS accuracy compared to \gqa, which proves that \ourdata highlights a novel reasoning task that can not be generalized using \gqa. Furthermore, the smaller performance gap among these models on \textsc{VisReas} suggests the inefficacy of the current {\vlm}s on our proposed spatial reasoning task.

\textbf{[\textcolor{purple}{Code-GEN}] From modular Code Generation models,} we analyze recent works - ViperGPT and VisProg. These models employ an \llm to generate an executable program that utilizes a pre-defined API, including functions such as \texttt{detect(image, obj\_category)} or \texttt{segment(image, obj\_category)}. VisProg also utilizes the ``in-context learning'' abilities of {\llm}s, enabling the model to respond to new queries with just a few examples of input and output behavior. Zero-shot evaluations of Code-GEN models on \gqa and \ourdata reveal that current models are struggling with our task more than \gqa, where both corpora use similar images. 
Specifically, their poor performance in \textit{Non-Problematic} questions denotes the inability of these models to cluster objects based on commonalities.

\subsection{Analysis}

According to \autoref{tab: baselines_breakdown}, all the models including \gpt struggle in \texttt{Compare}, \texttt{Count}, and \texttt{Query} question-types which require grounding, clustering, and verifying the existence of multiple objects, relations, and attributes. Specifically in \texttt{Query}, the performance gap between humans and the models is significantly higher which demonstrates the limitation of current models to perform complex multi-hop reasoning. \ours, on the other hand, shows a promising result in \texttt{Query} questions. We hypothesize that structured pseudocode helps the model consider each object and its corresponding attributes and relations before answering while the other models try to learn from the surface-level word distribution. In addition, \texttt{Query} questions are in general lengthier than other types of questions which makes it easier for the models to lose attention to the details (\autoref{fig: cat_length}).


\def\rot{\rotatebox}



In contrast, \gpt outperforms all generative models in \texttt{Problematic} questions. After analyzing the predictions, we find that \gpt excels at identifying problematic questions that involve an object not present in the image or an object with a false attribute. However, when the question becomes problematic due to an incorrect relation, \gpt consistently struggles to recognize it which also holds for other models. This signifies the uniqueness of our corpus that emphasizes understanding relations beyond simple object detection. 

To investigate the effect of \llm's scale on the \vqa task, we test two versions of {\llm}s (\vicuna 7B and 13B) within \ourdata architecture. \autoref{tab:7B-13B} breaks down the performance of \ours in the presence of different {\llm}s. We observe that increasing {\llm}'s size dramatically increases the accuracy of longer questions (\autoref{fig: cat_length}) such as \texttt{Non-Problematic}, \texttt{Count}, \texttt{Query}, and \texttt{Compare} instances and marginally improves performance on question categories such as \texttt{Choose} and \texttt{Verify}. This finding reassures the ability of larger \llm to reason with longer context. However, for problematic questions, increasing \llm size has no impact. As this category requires verification and grounding of information with image, both \llm and vision-language alignment need to be strong to excel in this domain.  




\begin{table}[t]
\centering
\resizebox{\columnwidth}{!}{%
%

\begin{tabular}{@{}lcccccccc@{}}
   Model
   & \rotatebox{60}{Choose}
   & \rotatebox{60}{Compare}
   & \rotatebox{60}{Count}
   & \rotatebox{60}{Query}
   & \rotatebox{60}{Verify}
   & \rotatebox{60}{Prob.}
   & \rotatebox{60}{Non-Prob.}
   & All\\
   \toprule
7B  & 81.20 & 54.90 & 35.13 & 59.24  & 82.75  & 55.38 & 63.92 & \textbf{62.74}\\
13B &  82.54 & 59.25 & 39.47 & 63.79 & 84.45 & 55.34 & 67.94 & \textbf{66.20} \\
\bottomrule
\end{tabular}

}
    \caption{Breakdown of accuracies on \textsc{VisReas} for \ours's \vicuna model size. We observe that \vicuna's model size improves performance in most question-types except the problematic ones.}
    \label{tab:7B-13B}
\end{table}

\section{Conclusion}

We introduce the \ourdata dataset, for real-world complex and multihop visual reasoning and compositional question answering. The dataset emphasizes object commonalities, differences, and relational aspects, necessitating validation of question-text relevance with the image before answering. We describe the dataset curation process along with the performance of SOTA models from three different domains in our task. Addressing the shortcomings in grounding and clustering in recent models, we propose a novel \ours baseline that deconstructs questions into pseudocodes and sequentially executes them using images to generate answers. We anticipate that this dataset and model will catalyze advancements in VQA research, pushing it toward complex semantic comprehension, robust reasoning, and addressing unanswerability when the provided context is not sufficient. 

\section{Discussion and Future Work}

Solving \vqa tasks via code generation and external APIs has gained attention due to its capability to perform complex reasoning and planning in a modular manner. However, code generation has limitations: a fixed set of operations limits models to specific types of questions and heavy use of external modules prevents end-to-end training. While modularity encourages specialization, in practice it requires managing multiple environments and heavy GPU memory usage as multiple large models are used to carry out visual and cognitive tasks like detection and captioning. In addition, current code generation methods \cite{surismenon2023vipergpt, Gupta_2023_CVPR} 
rely on OpenAI's API to generate executable code which hinders the accessibility of benchmarking due to its high costs\footnote{Evaluation with VisProg requires approximately 2,500 tokens per question including in-context examples, prompts, and outputs. Using original \texttt{text-davinci-003} model used in original code would cost $(0.0200 / \text{1000 tokens}) \cdot 2500~\text{tokens} \cdot 17171~\text{instances} \approx 858~\text{USD}$.} and fluctuations of OpenAI models over time \cite{chen2023chatgpts} 
which makes it hard to diagnose whether certain performance gains come from OpenAI model or improvements in other components. In contrast, our model and dataset suggest that one can use a single \vlm model that combines both the strength of structured reasoning and train it in a simple end-to-end manner. 
\ourdata requires many operations such as \texttt{select}, \texttt{filter}, \texttt{relate}, and \texttt{query} which are limited to cognitive skills to standard \vqa tasks and spatial reasoning. Therefore, models trained on \ourdata may not generalize well for visual-language tasks such as visual storytelling and image captioning which goes beyond the scope of our dataset. A natural future direction would be to incorporate other visual-language tasks into the dataset as well.

\bibliography{acl2023}
\bibliographystyle{acl_natbib}

\appendix



\section{Data Balancing} 
\label{appendix: data_balance}
A primary concern with current VQA datasets is the prevalence of question-conditional biases, enabling models to make informed guesses without a genuine grasp of the underlying images. Nevertheless, precise rendering of question semantics could offer enhanced control over these biases, holding the potential to significantly mitigate the issue \cite{zhang2016yin, kafle2017visual}. Motivated by this observation, we perform a rigorous balancing based on question categories, attribute/relation types, and answer distribution. 

Adopting the balancing approach outlined in previous research \cite{hudson2018gqa}, we employ a clustering strategy based on a fusion of two labels: \texttt{<attr/rel\_type>} and \texttt{<res\_type>}. The former denotes attributes or relation names (e.g., \textit{red} or \textit{right}), while the latter signifies reasoning types (e.g., \textit{verify.rel}). We refine the question set within each cluster, filtering out questions that encompass overlapping sets of objects in their texts or that contain subsets of objects already covered by other questions with complete sets. We prioritize questions featuring larger sets of objects and multihop relations, provided their length stays below 25. Finally, we introduce an additional label \texttt{<answer>} and equilibrate the question sampling through the answer distribution. After executing this balancing in an iterative manner on 2.07M questions, we generate a balanced corpus of 72,244 questions with images.

\section{Overview and Analysis of the \textsc{VisReas}}

This section provides an in-depth examination of the \textsc{VisReas} dataset, focusing on various aspects of question types and their characteristics. It encompasses an overview of question types, the distribution of semantic lengths, question readability scores, average question lengths per reasoning type, the relationship between question frequency and the number of attributes, and human accuracy on attributed questions.

\subsection{Questions Types and Templates}

\begin{figure*}[h!]
    \centering
    \includegraphics[width=\textwidth]{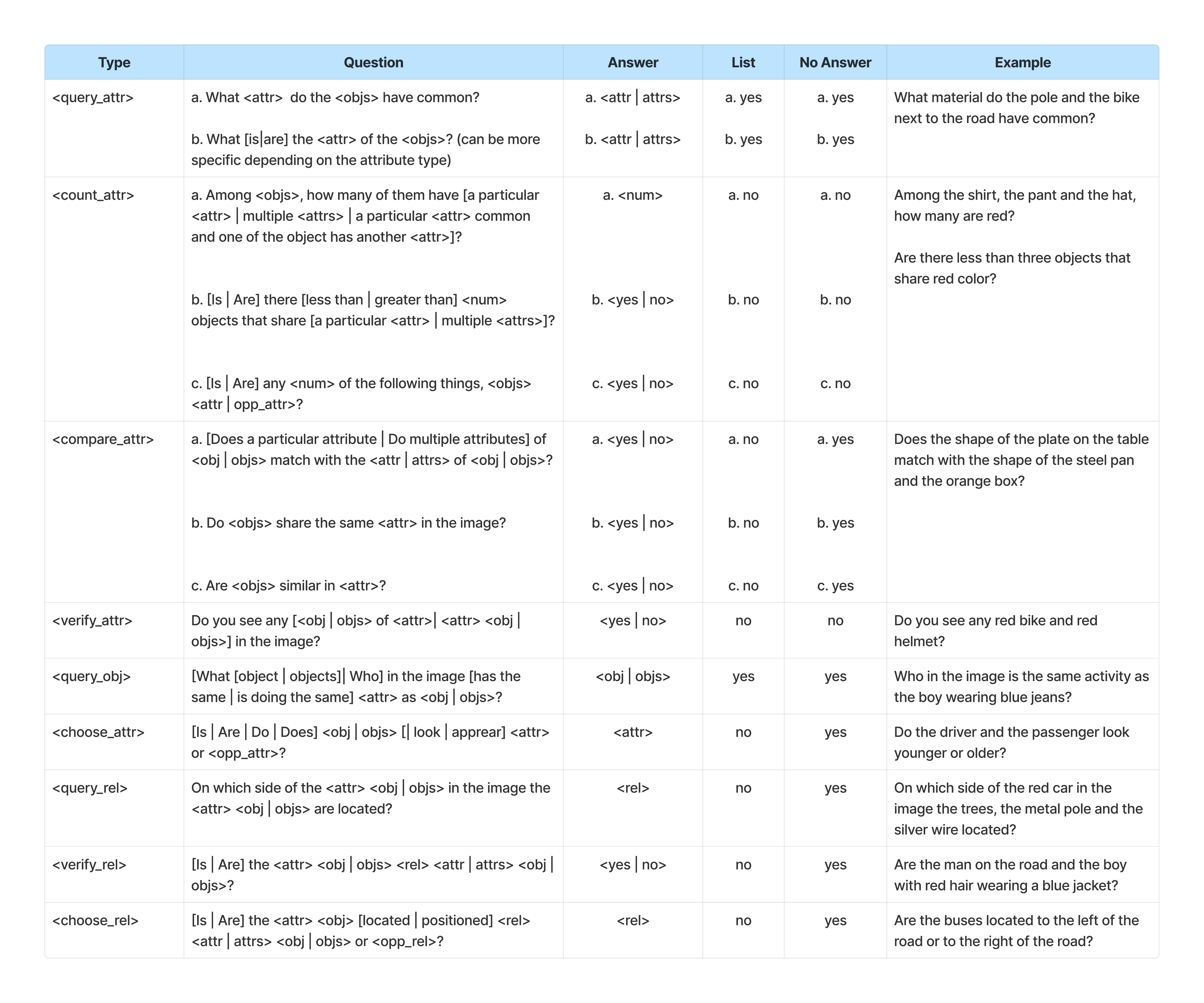}
    \caption{Overview of types of questions along with some templates and examples from the \textsc{VisReas} corpus.}
    \label{fig:ques_type}
\end{figure*}

The \textsc{VisReas} dataset features a diverse array of question types that challenge multimodal reasoning and compositional understanding. These question types include \texttt{query}, \texttt{count}, \texttt{compare}, \texttt{verify}, and \texttt{choose}, each requiring a unique approach to answer. Depending on how the clusters are made, each question type can further be broken down into \texttt{attr} and \texttt{rel} subtypes. Therefore, in total, there can be nine categories of questions. Figure \ref{fig:ques_type} gathers all templates and examples from the dataset to offer insights into the intricacies of these question categories.

\subsection{Distribution of Relation Hops and Readability}

\begin{figure}[h]
    \centering
    \includegraphics[width=0.45\textwidth]{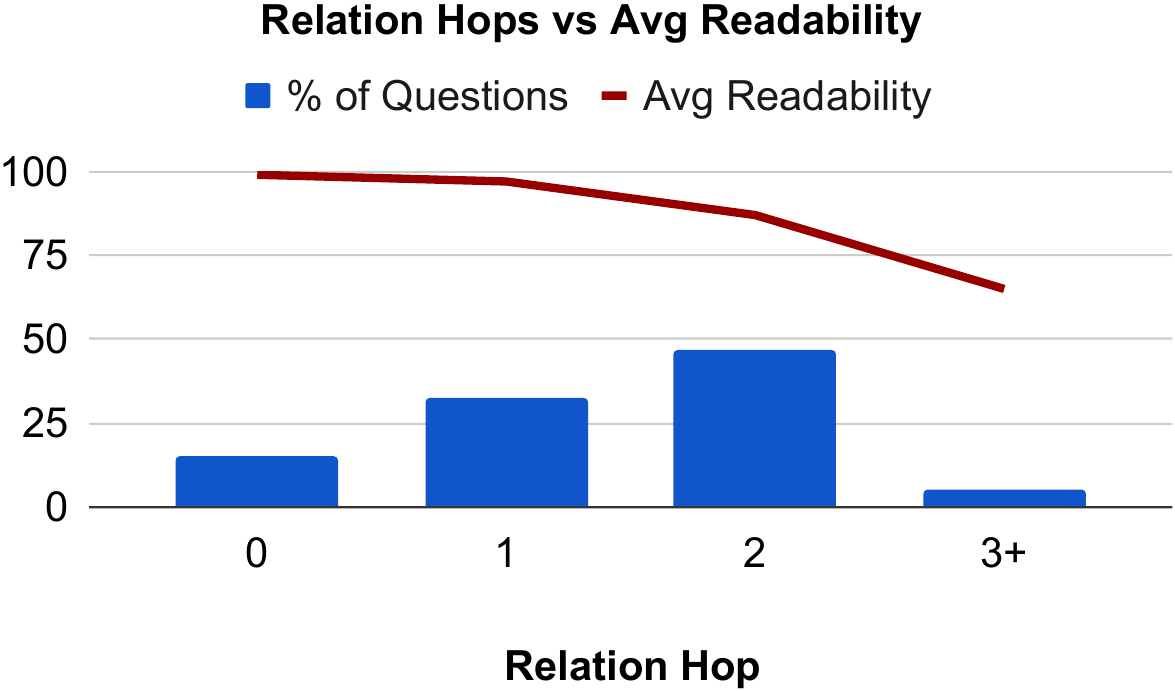}
    \caption{Distribution of \textsc{VisReas} questions semantic length (number of computation steps to arrive at the answer) as well as the readability scores for each semantic step type. We can see that most questions require at most two reasoning steps, where each step may involve tracking a relation between objects, an attribute identification, or a logical operation. At the same time, questions with larger semantic steps are difficult to read.}
    \label{fig: readability}
\end{figure}

A comprehensive analysis of the distribution of relation hops in \textsc{VisReas} questions reveals a predominant trend toward questions that involve about two reasoning hops. These hops can entail tracking object relations, identifying attributes, or executing logical operations. We conduct a readability test using the workers from Amazon Mechanical Turk. Our analysis reveals that questions with larger relation hops demonstrate a noticeable decline in readability, emphasizing the complexity associated with extended reasoning (Figure \ref{fig: readability}). To enhance the quality of the dataset so that it can reflect the real-world day-to-day life questions, we choose to keep the relation hop within two.

\subsection{Average Question Length per Reasoning Type}

\begin{figure}
\centering
\begin{subfigure}{.4\columnwidth}
  \centering  \includegraphics[width=0.7\linewidth]{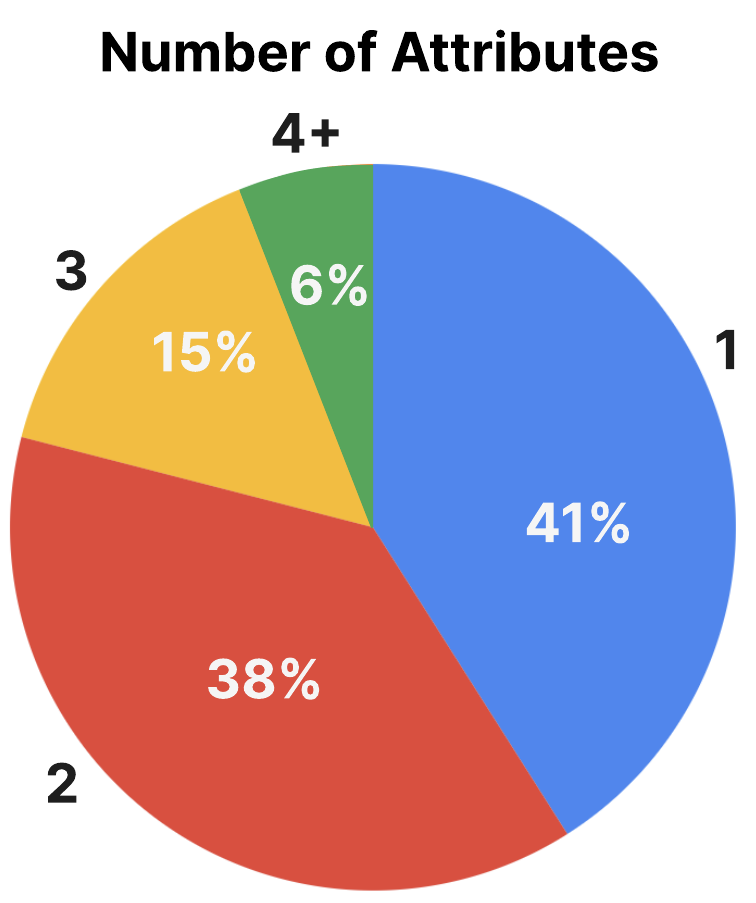}
  \caption{}
  \label{fig: ques_n_attributes}
\end{subfigure}%
\begin{subfigure}{.6\columnwidth}
  \centering
  \includegraphics[width=\linewidth]{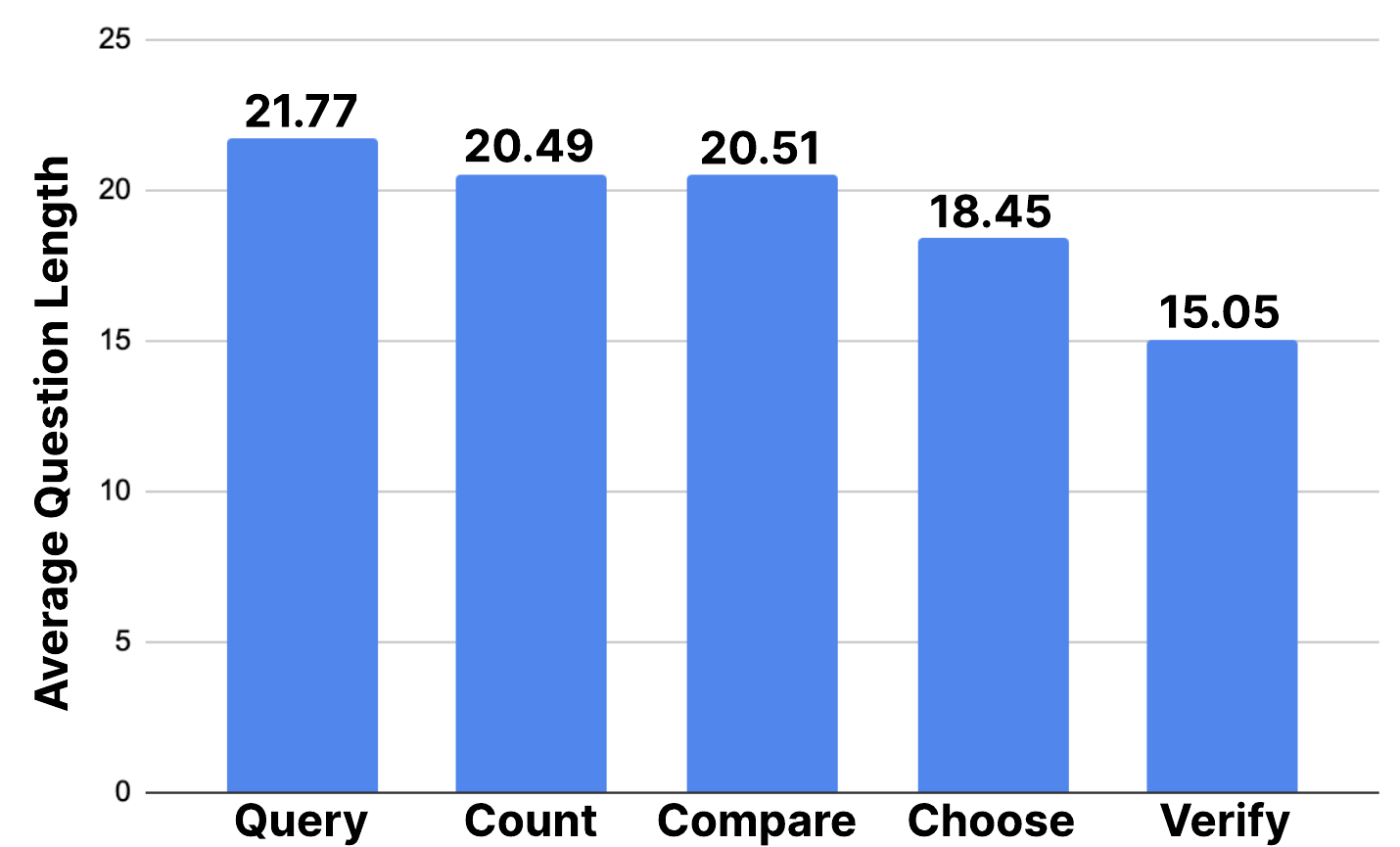}
  \caption{}
  \label{fig: cat_length}
\end{subfigure}
\caption{(a) Question distribution across the number of attributes in a query. The question complexity increases with the number of attributes or relations. (b) Average question length per reasoning type in \ourdata corpus. Query questions are lengthier than other reasoning categories as these questions contain multiple objects of similar attributes with their relations.}
\label{fig:q_features}
\end{figure}

By dissecting question lengths across different reasoning categories in Figure \ref{fig: cat_length}, we observe a consistent trend: query questions tend to be longer than other reasoning types. This phenomenon is particularly apparent due to the inclusion of multiple objects sharing similar attributes and their corresponding relations.

\begin{figure}[h]
    \centering
    \includegraphics[width=\columnwidth, height=25mm]{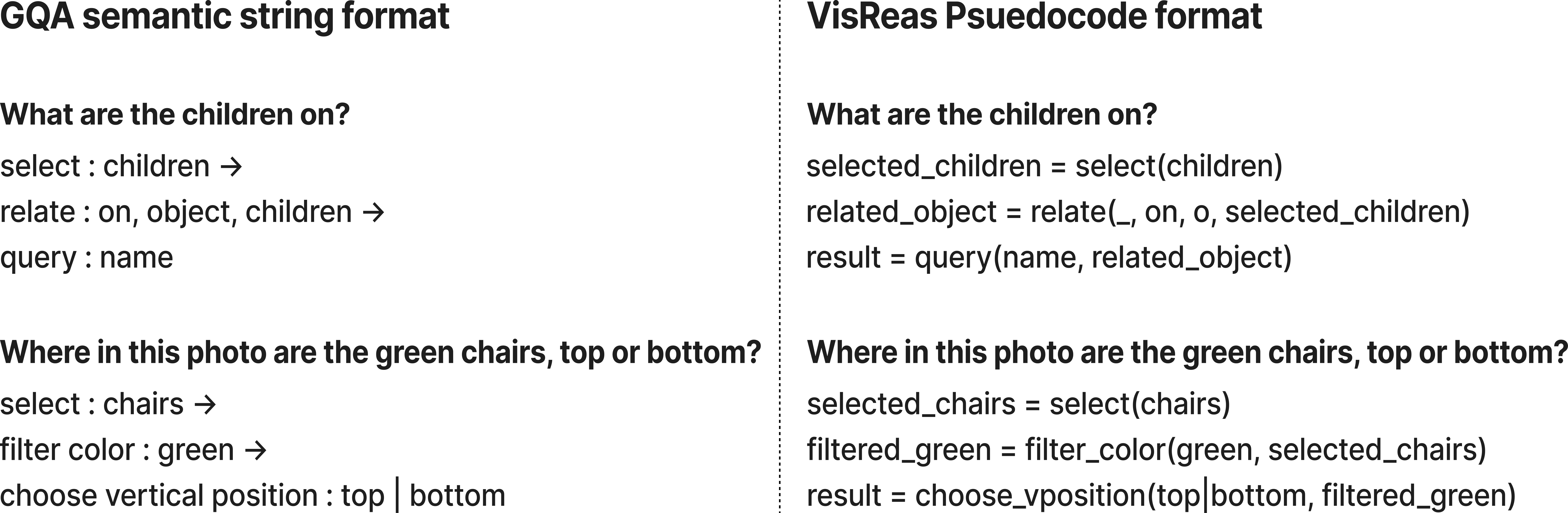}
    \caption{\textbf{Pseudocode format.} Our method re-structures the format of \gqa semantic string to pseudocode to better leverage Code-LLMs without adding any auxiliary information.}
    \label{fig: pseudocode}
\end{figure}

\begin{figure}[h]
    \centering
    \includegraphics[width=\columnwidth, height=60mm]{images/semantic-parser.pdf}
    \caption{\textbf{Semantic string parser.} For every line of semantic string, we use regex and string manipulation to extract operator and its arguments. We represent scene-graph in adjacency list format and run the parsed operator to get formatted pseudocode and its expected output.}
    \label{fig: semantic_parser}
\end{figure}

\begin{figure*}[h!]
    \centering
    \includegraphics[width=\textwidth]{images/star-chain-code.pdf}
    \caption{Overview of pseudocodes for two different traversal types in the \textsc{VisReas} corpus.}
    \label{fig: star-chain-code}
\end{figure*}





\subsection{Question Frequency and Attribute Usage}

The \textsc{VisReas} corpus has been generated using the clusters of objects that share similar relation or attribute. However, clusters based on shared attributes/relations can share objects that possess all of those attributes/relations. For example, a table and a chair have the \texttt{color} \textit{brown} and \texttt{material} \textit{wood} in an image. Initially, we have two clusters with \textit{brown} and \textit{wood}. Now, if both clusters share some objects, we again create a new cluster based on \textit{brown}+\textit{wood} adding the shared objects (i.e., table and chair). Using this approach, we create clusters that share multiple attributes and relations and generate questions that involve filtering multiple attributes/relations along with the identification of objects of interest. Figure \ref{fig: ques_n_attributes} shows the distribution of questions in \textsc{VisReas} with respect to the number of attributes/relations. As the number of attributes/relations goes higher, the number of clusters also decreases resulting in decreasing number of questions.


    

\subsection{Human Accuracy on Attributed Questions}

\begin{figure}[h]
    \centering
    \includegraphics[width=\columnwidth]{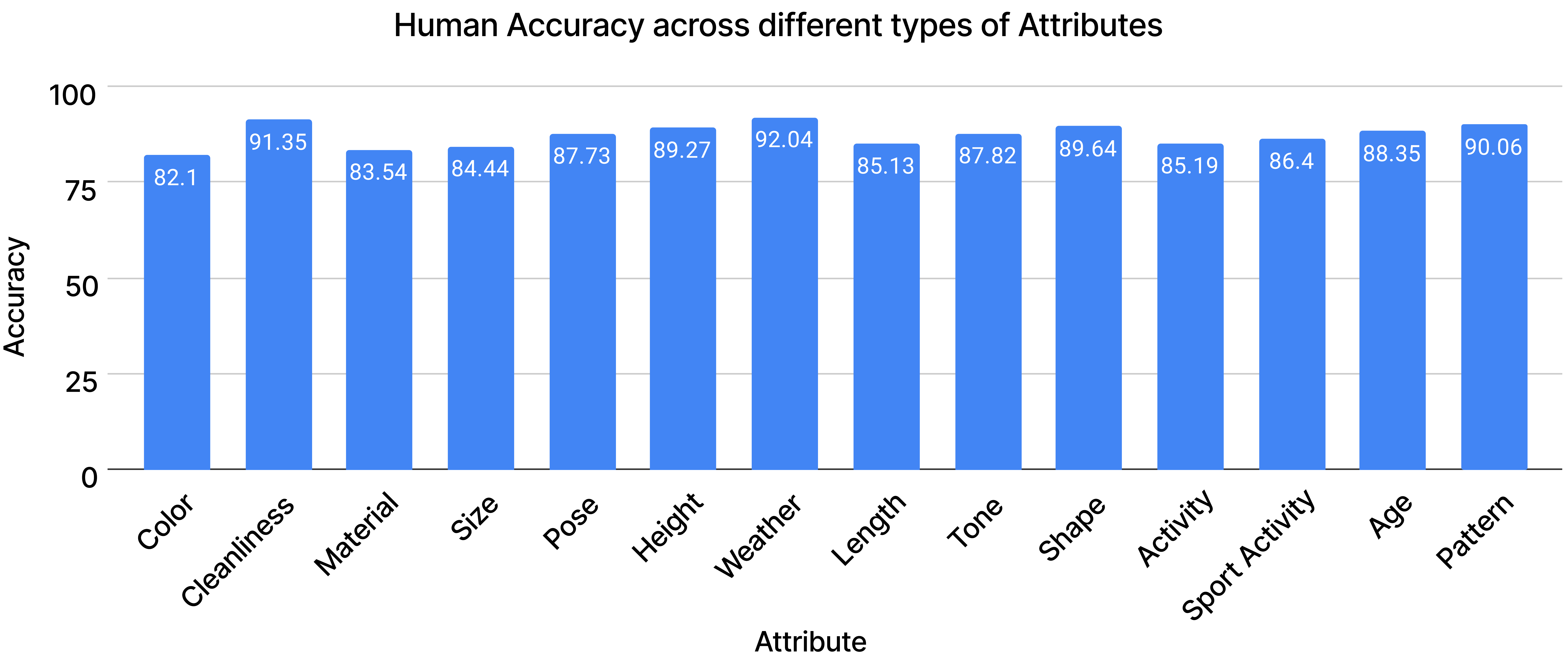}
    \caption{Human accuracy on different attributed questions}
    \label{fig: attr_hum_acc}
\end{figure}

The final facet of our exploration delves into human accuracy when answering attributed questions from the \textsc{VisReas} dataset. By assessing the performance of human subjects across different question types and attributes, we gain a deeper understanding of the challenges inherent to this multimodal reasoning task. Figure \ref{fig: attr_hum_acc} breaks down the human accuracy across different attribute types. It is noticeable that \texttt{color} and \texttt{material} questions have the lowest accuracy, as they contain a higher amount of questions compared to other attributes.

In summary, this section offers a comprehensive overview and analysis of the \textsc{VisReas} dataset, encompassing question types, semantic lengths, question readability, average lengths per reasoning type, attribute-based question distribution, and human accuracy. These insights contribute to a holistic understanding of the dataset's intricacies and its potential to advance the field of visual reasoning and question answering.

\section{Baseline Configuration}\label{appendix: baseline_config}

All baselines follow default settings provided by the original author evaluation script. All configurations for model, optimizer, scheduler, and training follow default parameters from Pytorch and Huggingface library. For generative models, all inference is done using default settings without temperature tuning, nucleus sampling, repetition penalty, etc. Specific settings used for zeroshot and finetuning are presented below:

\subsection{VisProg}

The original VisProg script uses \texttt{text-davinci-003} model which is around 10 times more expensive than \texttt{gpt-3.5-turbo} model. To cut evaluation costs, we use the \texttt{gpt-3.5-turbo} model instead. All 20 examples are found in \gqa evaluation script for code generation.

\subsection{ViperGPT}

For similar reason as VisProg, we use \texttt{gpt-3.5-turbo} model for code generation to reduce costs. Since generated code doesn't always return functional Python code, we return either ``\texttt{None}'' or ``\texttt{ERROR}'' in these cases. In cases where the code throws an error, the answer defaults to ``\texttt{ERROR}''. In cases where the code didn't have a return statement, the answer defaults to ``\texttt{None}''.

\subsection{\ours}
\label{subsec: code2vision}



The effective batch size is kept at 4 across experiments. LoRA modules are only attached to query and value linear layers in attention layers. The batch size and gradient accumulation steps are adjusted accordingly. Due to memory requirements, we set batch size to 1 on each GPU and set gradient accumulation steps to 4. We've have used 2-4 A6000 GPUs with distributed data parallel (DDP) strategy for multi-GPU training. Training \ours on \textsc{VisReas} takes around 13 hours using 2 A6000 with LLaVA-1.5 backbone.

\begin{table}[h!]
\centering
\resizebox{0.9\columnwidth}{!}{%
\begin{tabular}{l|c}
\hline
\textbf{Hyperparameters} & \textbf{Values}                 \\ \hline
Effective batch size     & 4                               \\
Learning rate            & 5e-6 (\gqa), 2e-5 (\textsc{VisReas})                            \\
Precision                & bfloat16                        \\
Optimiser                & AdamW                           \\
Schedule                 & Linear warmup with cosine decay \\
Warmup steps             & 128                             \\
Epoch                    & 1                               \\ \hline
\end{tabular}%
}
\caption{Hyperparameters for \ours model}
\end{table}

\begin{table}[h!]
\centering
\resizebox{0.5\columnwidth}{!}{%
\begin{tabular}{l|c}
\hline
\textbf{Hyperparameters} & \textbf{Values} \\ \hline
Rank            & 8               \\
Alpha                    & 16              \\
Dropout                  & 0.05            \\ \hline
\end{tabular}%
}
\caption{LoRA configurations}
\end{table}

\subsection{InstructBLIP / BLIP-2 / LLaVA-1.5}
\label{subsec: gen_configs}

On \gqa, we use identical configuration as \ours for LLaVA-1.5. For InstructBLIP and BLIP-2, we observe that batch size of 4 causes the model to output repetitive tokens during inference. For that reason, we increase the effective batch size to 8. We use the same original prompt that the authors have reported in their original papers.

On \textsc{VisReas}, we again use identical configuration as \ours for LLaVA-1.5. For InstructBLIP and BLIP-2, we lower the learning rate to 5e-6 and increase the effective batch size to 8 for the same reason above.





\subsection{LXMERT / ViLBERT / CRF}

For all three models trained with the classification task, we used the default hyperparameters that have been used to finetune on GQA corpus for consistency. As GQA and \textsc{VisReas} share the same image and scenegraphs, using the same model with the same configuration should produce different results if the two tasks are different. And the result section reflects the distinction between GQA and \textsc{VisReas}.
\begin{table}[h!]
\centering
\resizebox{\columnwidth}{!}{%
\begin{tabular}{l|ccc}
\hline
\textbf{Hyperparameters} & \textbf{LXMERT} & \textbf{ViLBERT} & \textbf{CRF}  \\ \hline
Learning rate            & 1e-5            & 0.00004          & 1e-4          \\
Optimizer                & BertAdam        & AdamW            & BertAdam      \\
Schedule                 & Linear Warmup   & Linear Warmup    & Linear Warmup \\
Epoch                    & 4               & 20               & 13            \\ \hline
\end{tabular}%
}
\caption{Hyperparameters of all CLS baselines}
\end{table}
 

\section{Effect of pseudocode finetuning}

We study the effect of finetuning a VLM to perform VQA through pseudocode-guided reasoning. Table \ref{tab:code_nocode} demonstrates that finetuning LLaVA-1.5 to follow pseudocode consistently improves performance on \textsc{VisReas} for both 7B and 13B models.

\begin{table}[h!]
    \centering
    \resizebox{\columnwidth}{!}{%
    \begin{tabular}{l|cc}
    \hline 
         Model size & \textbf{Without Pseudocode} & \textbf{With Pseudocode} \\ \hline
        7B & 57.36 & 62.74 \\
        13B & 61.38 & 66.20 \\ \hline 
    \end{tabular}
    }
    \caption{Effect of pseucode finetuning on LLaVA-1.5}
    \label{tab:code_nocode}
\end{table}

\section{Examples from \textsc{VisReas} and GQA}

\begin{figure*}[h!]
    \centering
    \includegraphics[width=\textwidth]{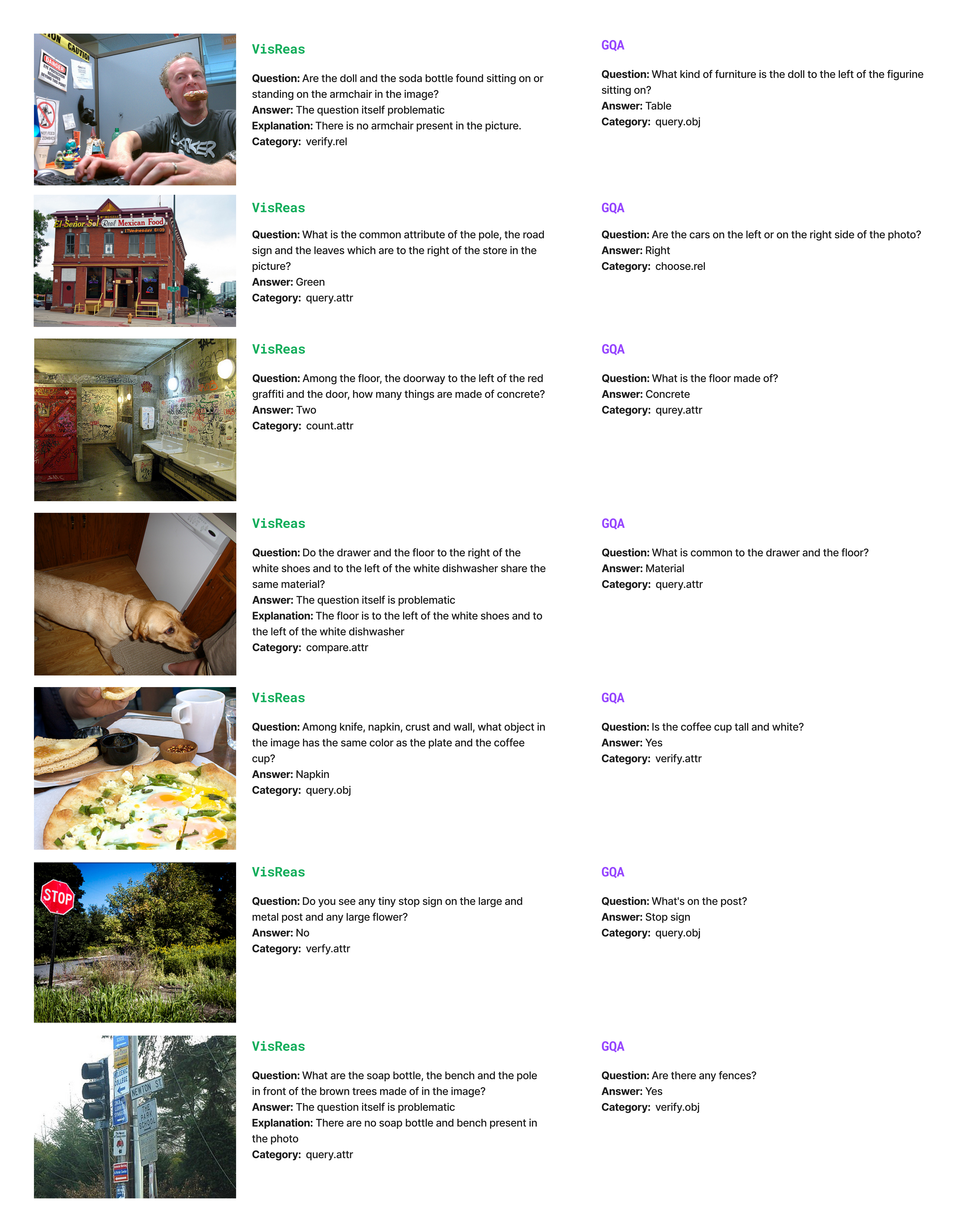}
    \caption{Example questions from the \textsc{VisReas} and the GQA corpuses.}
    \label{fig:ques_example}
\end{figure*}

In Figure \ref{fig:ques_example}, we show example questions from \textsc{VisReas} and GQA using the same image. In general, \textsc{VisReas} tends to have longer questions compared to GQA. Additionally, \textsc{VisReas} questions involve more than two objects, whereas GQA primarily centers on one or two objects.

\section{Mechanical Turk Details}

To evaluate human performance, we used Amazon Mechanical Turk to collect human responses for 5000 random questions, taking a majority vote among three workers for each question. We limited our pool of crowdworkers to individuals located in the US or Canada, requiring a minimum of 1,000 previously approved HITs with a 95\% approval rate. Additionally, participants had to achieve a minimum score of 70\% or higher on our qualification task before gaining access to our main task. In the subsequent sections, we provide details of this response collection process.

\subsection{Qualification Test for Worker Selection}

To secure accurate human assessments, we carefully designed a qualification test using Amazon Mechanical Turk interfaces (Figure \ref{fig: mturk1}). This test aimed to select proficient workers capable of accurately completing the \textsc{VisReas} task: (1) The qualification test encompassed two distinct tasks. The initial task focused on careful comprehension of instructions. Workers were required to attentively read the instructions and subsequently answer a set of multiple-choice questions to assess their grasp of the task's nuances. (2) Upon successful completion of the first task, the qualified workers proceeded to the task proficiency evaluation stage. Here, a series of ten questions, each accompanied by an image, were presented. The workers' task was to select the correct answer from a dropdown list of 2013 entries. The selection process for the final evaluation cohort prioritized workers who achieved correct answers for more than seven out of the ten questions. 

\begin{figure*}[h]
    \centering
    \includegraphics[width=2\columnwidth]{images/mturk_whole.pdf}
    \caption{Amazon Mechanical Turk interfaces used for Qualification Test to choose the right workers for human accuracy assessment on \textsc{VisReas} task. We study the workers by deploying two tasks. In the first task, we ask the workers to read the instructions carefully (\textbf{Top left}) and answer some multiple-choice questions (\textbf{Top right}). After passing this task, ten questions with images will be presented and the final task would be to choose the right answer from the answer dropdown list (\textbf{Bottom right}). We choose the workers for the final evaluation who have correctly predicted more than seven answers out of ten questions.}
    \label{fig: mturk1}
\end{figure*}

\subsection{Human Accuracy Assessment Interfaces}

After gathering qualified workers who are aware and proficient in our task, we move to the final stage of the evaluation process (Figure \ref{fig: mturk_HIT}). For each Human Intelligence Task (HIT), an image and the corresponding question were provided. Workers were tasked with selecting the correct answer from the same dropdown list used for the worker selection stage. Furthermore, we requested workers to rate the complexity and structural integrity of the presented question, thereby acquiring insights into the inherent challenges posed by various question types.

To facilitate a deeper understanding of the potential issues with the queries, we encouraged workers to provide additional details about any perceived problems. If a worker identified a problematic aspect within the question, they were encouraged to rephrase or rewrite the query to address the issue. This dynamic engagement aimed to uncover underlying complexities and refine the evaluation process.

\begin{figure*}[h]
    \centering
    \begin{subfigure}[b]{2\columnwidth}
         \centering
         \includegraphics[width=\columnwidth]{images/mturk_final_q.pdf}
        \caption{}
        \label{fig: mturk2}
     \end{subfigure}
     \vfill
     \begin{subfigure}[b]{2\columnwidth}
         \centering
         \includegraphics[width=1\columnwidth]{images/mturk_yes_no.pdf}
        \caption{}
        \label{fig: mturk3}
     \end{subfigure}
     \caption{Amazon Mechanical Turk interfaces for human accuracy assessment on \textsc{VisReas} task using the qualified workers. (a) For each HIT, we provide an image and a question that needs to be answered from a dropdown list of 2013 entries. In addition, we ask for rating the complexity and structural soundness of the query and further look for details if any Turker finds the question problematic. (b) To investigate what type of problem the question possesses, we ask for further details from the workers and even encourage them to rewrite the query to remove the problem they faced while answering the query.}
    \label{fig: mturk_HIT}
\end{figure*}

\end{document}